\documentclass{article}

\usepackage[final]{nips_2016}
\usepackage{graphicx}
\usepackage[utf8]{inputenc} 
\usepackage[T1]{fontenc}    
\usepackage{hyperref}       
\usepackage{url}            
\usepackage{booktabs}       
\usepackage{amsfonts}       
\usepackage{nicefrac}       
\usepackage{microtype}      

\title{Mapping chemical performance on molecular structures using locally interpretable explanations}

\author{
  Leanne S.~Whitmore, Anthe George, Corey M. Hudson\thanks{LW,AG,CH are also affiliates at the Joint BioEnergy Institute, Emeryville, CA, USA 94608.}\\
  Sandia National Laboratory\\
  Livermore, CA, USA 94551\\
  \texttt{\{lwhitmo,ageorge,cmhudso\}@sandia.gov} \\
}

\begin{document}

\maketitle
\begin{abstract}
In this work, we present an application of Locally Interpretable Machine-Agnostic Explanations to 2-D chemical structures. Using this framework we are able to provide a structural interpretation for an existing black-box model for classifying biologically produced fuel compounds with regard to Research Octane Number. This method of "painting" locally interpretable explanations onto 2-D chemical structures replicates the chemical intuition of synthetic chemists, allowing researchers in the field to directly accept, reject, inform and evaluate decisions underlying inscrutably complex quantitative structure-activity relationship models.
\end{abstract}

\section{Introduction}

The number of known organic molecules (what is referred to as the {\it known chemical space}) is on the order of $10^6-10^7$ reported structures [1]. This is considerably smaller than the potential structural space of pharmaceutical chemicals (of the order $10^{60}$ [2]), and which itself only represents a fraction of potentially important chemicals. This combinatorial explosion of potential chemical structures is the reason why the rational design of a chemical with a desired set of chemical properties is so challenging. First-chemical-principles screening methods (like molecular dynamics simulation and density functional theory), which apply the list of known atomic interactions and then determine the value of a desired chemical property are time consuming (taking between hours and days per molecule, a much longer amount of time to prepare data for, and requiring super-computing resources). Nonetheless, these techniques are able to directly bridge the gap between what is known structurally about a molecule and the measurable function of interest. Machine-learning methods, which are typically fast to implement and highly accurate, often fail to provide a satisfying mapping between molecular structure and the desired property of interest. This gap in interpretation limits the amount of knowledge that can be gained using standard machine-learning methods. When these models fail, it is unclear why. And when they succeed, it is unclear how to tie these successes to a widening chemical intuition. The lack of structural interpretability and molecular explanability has resulted in distrust of {\it black-box} methods. This distrust ultimately led to the Environment Directorate of the Organisation for Economic and Co-Operation and Development mandating a call for mechanistic bases for machine learned quantitative structural activity relationships [9].

\subsection{Motivation for studying the Research Octane Number}

Research Octane Number (RON) is an important chemical property. In spark ignition (SI) engines, RON measures the propensity of a fuel to avoid compressive pre-ignition in an engine prior to sparking. In conventional SI engines low-RON fuels produce "knock", where fuels ignite prior to spark, which leads to low performance and engine damage. In advanced engines where fuel blending is more tightly managed, the issue is even more pronounced, and low-RON fuels lead to a phenomenon known as "super-knock" which undermines performance and can seriously damage the engine [10]. To make biologically produced chemicals that offset greenhouse gas producing petrochemicals in a fuel blend commercially viable, it is necessary that they improve the performance of the fuel into which they are blended [11].

\section{Research Octane Number Machine Learning Classifier}

Whitmore et al. (2016) recently developed a RON classifier, using Random Forest classification, and applied it to a variety of biologically produced compounds (hereafter bcmlRON). This classifier was built using easily collected chemical properties for 148 training compounds. This model allows any biologically produced hydrocarbon/oxygenate compound to be rapidly classified into high or low RON (defined at RON = 94.4). In 100 random 50\% leave-out experiments this model had a mean accuracy of 87.79\% (standard deviation of 5.8\%), a mean precision of 88.16\% (standard deviation of 11.4\%) and mean recall of 88.17\% (standard deviation of 11.3\%) with a Receiver Operator Characteristic AUC of 0.88 (standard deviation of 0.06). Experimental validation of this model was performed using 16 compounds that had the RON directly measured in the laboratory. The accuracy of this model was very high, and found to be between 93.75\% and 100.0\% depending on the confidence interval around the decision boundary.

Despite the success of the bcmlRON model in predicting RON in unknown compounds, the interpretability of this model in physical/structural space is limited. Following feature reduction, the only structural features that contribute to bcmlRON's classification is the presence/absence of 5 and 6 carbon bonds (C-C-C-C-C and C-C-C-C-C-C in the {\it standard simplified molecular-input line entry system}, SMILES). The other features (i.e., XLogP3, auto-ignition, LogP, XLogP-AA, complexity, boiling point and vapor pressure) are functions of chemistry and thermodynamics. However, an essential hypothesis in chemistry is that the underlying basis of chemical property prediction is fundamentally structural and chemical properties are interpretable as extensions of first-principle structural properties. The features used in bcmlRON classifier may provide some insight into the factors underlying RON, but from a synthetic chemistry perspective, most of the properties are as intuitive and difficult to engineer in a compound and RON.

\section{Tying bcmlRON to locally interpretable chemical structure}

\subsection{Background on LIME}
Ribeiro et al's (2016) framework for Locally Interpretable Model-Agnostic Explanations (LIME) provides a technique for extracting local explanations of features in {\it black-box} models. LIME works by perturbing $x$ where $x \in{\mathbb{R}^{d}}$ describes the feature vector of a particular instance and $x^\prime \in\{0,1\}^{d^{\prime}}$ is the binary vector for this representation. In this tool, $G$ is the total set of interpretable models, where $g\in G$ is a particular interpretable model. $\Omega (g)$ is the complexity of this model. $f(x)$ is a particular classifier. The measure of unfaithfulness of $g$ in approximating $f$ is defined as $\cal{L}$$(f,g,\Pi_x)$ where $\Pi_x(z)$ is the proximity mapping of $z$ to $x$. LIME works by minimizing $\cal{L}$$(f,g,\Pi_x)$, while keeping $\Omega (g)$ low enough to maintain human interpretability. This is defined as: 
\begin{equation}
\xi(x) = argmin_{g \in G} {\cal L }(f,g,\Pi_x) + \Omega (g)
\end{equation}

\subsection{Value of local interpretations for understanding RON}
Many of the chemical features used to build bcmlRON's Random Forest Classifier are sufficiently complex that they warrant their own explanatory models. In fact, cheminformaticists have worked to develop explanatory machine learning models for each of the chemical features used to build this model (i.e., XLogP3, auto-ignition, LogP, XLogP-AA, complexity, boiling point and vapor pressure) [5-8]. Therefore, more thoroughly complete chemical explanation would be able to explain RON using strictly atoms and atomic bonds. However, in doing so, the risk would be that the interrelationships between structural features would be too complex to be human understandable. Experienced fuel chemists develop an intuition for structural-property relationships, but this may take a career to develop and may still be difficult to interpret for chemical properties as complex and multi-faceted as RON.

\section{Mapping local structural interpretations onto 2-D chemical structures}

To map local structural property interpretations back onto global molecular structure we rebuilt bcmlRON using exclusively sub-structural properties. Originally this study measured 881 sub-structural fingerprints from the National Institute of Health's National Center for Biotechnology Information's PubChem resource [1]. Of these original features, only 131 were variable, and therefore potentially informative. The new bcmlRON model excludes complex chemical features, since they may be difficult to interpret. 

The PubChem sub-structural fingerprints can be viewed as traversal paths through chemical structures. Many of these include wild card, negation and iteration elements, much like a regular expression. For example, the symbol {\bf -} represents a single bond and {\bf C} represents a carbon atom. Therefore, C-C-C-C-C is a path that traverses five single-bonded carbon atoms. A more complex pattern is [a;!\#6]1aaaa1. In this case {\bf a} is an aromatic atom (meaning the sub-structure is ring-like and has a planar and stable ring of resonance bonds), {\bf \#6} is also a carbon (referring this time to its atomic number), {\bf !} is the symbol for negation, and a 1 at the beginning and end of a series signifies a ring structure. Therefore this sub-structure represents a path through a five atom aromatic ring, where one atom is not carbon. These features are binary variables, representing presence/absence of the sub-structure within the given molecule.

Rather than using LIME, one option for evaluating local importance is using feature importances, which can be estimated directly from the Random Forest Classifier. This provides a rather direct measure of local interpretability. The problem with including this in a structural interpretation of local feature weightings, is that these feature importances are gleaned through average changes in classification efficiency. These may take the form of average decreases in Gini impurity or the decrease in the accuracy, using out-of-bag observations [12]. Regardless of the means of calculating feature importances, they are fundamentally system-level, and do not provide directionality in their local importance. This is a problem, because it confounds the relative merit of including or excluding the feature. This reduces our ability to evaluate individual bonds. A feature may be highly important, but fall have variable directionality on a decision tree.

\subsection{Performance of the rebuilt bcmlRON}

The accuracy, precision, recall and ROC AUC of bcmlRON decreased using strictly structural features in the Random Forest Classifier. The performance is still high, after 100 random 50\% leave-out samples, the mean accuracy, precision, recall and ROC AUC are 83.41\% (+/- 3.61), 85.15\% (+/- 5.42), 81.83\% (+/- 7.07) and 0.84 (+/- 0.04) respectively.  There are a variety of reasons for the drop in performance, but the most likely reason relates to the fact that the complex features which were excluded in this classifier, mask a variety of subtle structural-functional relationships. A simple t-test between the two models, shows that this decrease in accuracy is statistically significant (t = 6.41, {\it P}<0.001).

\subsection{Local interpretations of sub-structural features}

\begin{figure}[ht]
  \includegraphics[width=\textwidth]{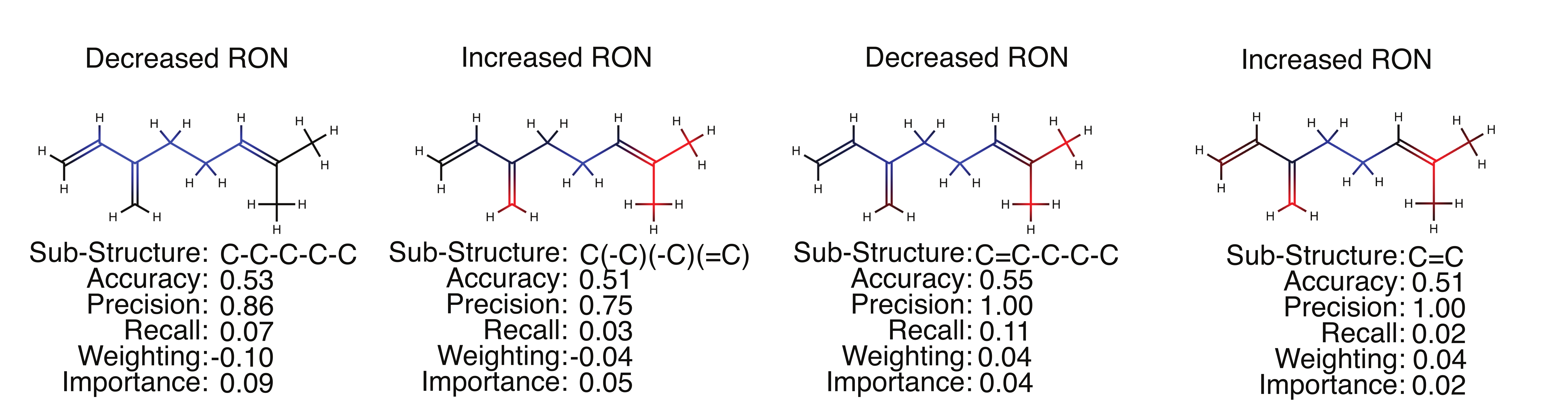}
  \caption{The 2-D structure of myrcene, a compound with a marginally high RON value of 82.5. In this case, myrcene has structural features that suggest both high and low RON. C-C-C-C-C and C=C-C-C-C are paths that are weighted toward decreased RON. C(-C)(-C)(=C) and C=C are sub-structural pathways that suggest bonds that lead to increased RON. Accuracy, precision and recall refer to the direct mapping between the binary feature and the classification of RON. Weighting is the local sub-structural weighting (using LIME). Importance is the average change in Gini Impurity.}
  \label{fig1}
\end{figure}

In this study we used the rebuilt bcmlRON model and LIME to determine locally interpretable weightings and painted these sub-structural features onto 2-D chemical representations. This was done for 16 compounds that were validated by experimental measurement. Figure \ref{fig1} provides an example of how these paintings function and also how they may be interpreted. We use LIME to extract locally interpretable weightings and "paint" these onto individual molecular sub-structures. The combination of these provides a total molecular coloring, where sub-structures that predict decreased RON are colored blue and sub-structures that predict increased RON are colored red. Overlapping sub-structures grade between these two extremes.

Since individual sub-structural features are binary variables, the individual accuracy, precision and recall of these can be evaluated independently. This is in keeping with the local interpretation approach, and gives us not only a sense of the locally interpretable weightings of these features in our model, but also their strict reliability. Using the original training set, the accuracy of sub-structural  features varies from 0.24 to 0.65, with a mean value of 0.51 and a relatively symmetric distribution with a variance 0.00. The precision of sub-structures varies from 0 to 1, but is high and highly skewed (mean = 0.75, variance = 0.10), especially relative to recall which is low and also highly skewed (mean = 0.14, variance = 0.04). As a classifier, no individual sub-structural feature is sufficient to satisfactorily explain the data at high accuracy, but nonetheless, cover much of the data's range of variability.

In the case of myrcene in Figure 1, it is clear that the weakest points for RON in this molecule are the central single bonds (colored in blue). This suggests that synthetic creation of double bonds in this region, or the removal of these single bonds would likely increase the RON well above the observed 82.5.

\begin{figure}[ht]
  \includegraphics[width=0.9\textwidth,keepaspectratio]{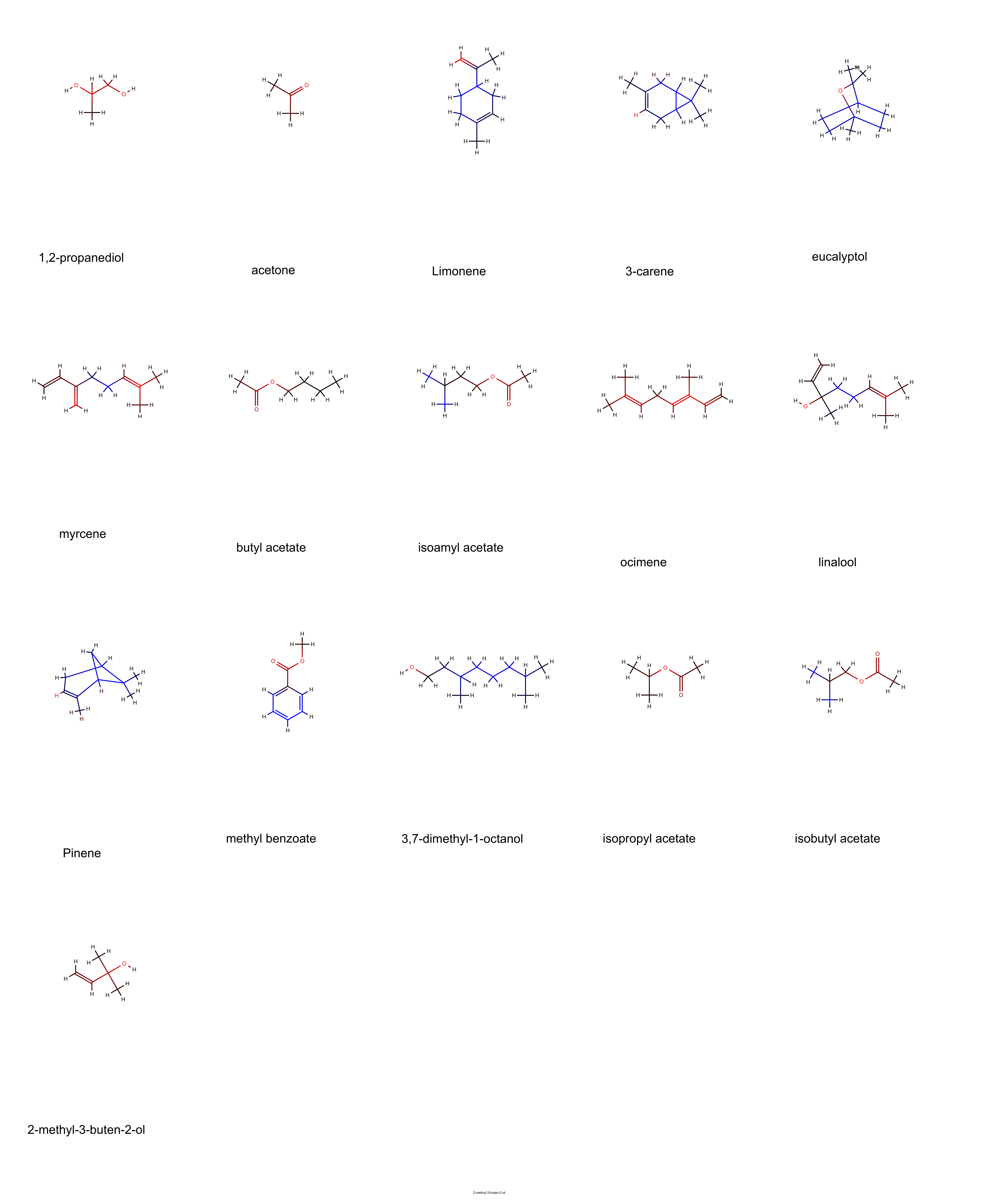}
  \caption{Sub-structural mapping of localized weightings for each of the 16 compounds that were experimentally validated. As with Figure 1, red indicates a sub-structural feature indicative of high RON and blue indicates a sub-structural feature indicative of low RON.}
  \label{fig2}
\end{figure}

In addition to the characterization of the single myrcene model, this study also presents the local sub-structural weightings for each of the 16 molecules that were previously validated. Figure \ref{fig2} provides a visualization of this comparative analysis. The ranked correlation between the average of tested local weightings (after 100 bootstraps) and the probability of high RON classification in the original bcmlRON is Spearman's $\rho$ = 0.784 \({\it P} < 0.01\). The  Spearman's correlation between the original bcmlRON and measured RON is $\rho$ = 0.761 \({\it P} < 0.01\). The new model is $\rho$ = 0.711 \({\it P} < 0.01\). This value is smaller, but not significantly different \({\it P} = 0.787\). The original bcmlRON model perfectly predicted the measured RON value. This locally-interpretable model performs well, but two particular compounds stand out: ocimene and eucalyptol. Ocimene has a measured RON of 72.9, but the new locally-interpretable model suggests a higher RON. One of the advantages of the LIME framework, is that it allows us to visualize where errors are occurring. The high presence of double carbon bonds suggest a high RON. Eucalyptol has a high measured RON (99.2), but the average local weighting is low (0.46). The large number of internal single carbon bonds suggests a low RON.

\section{Conclusion}

The advantage of this model is that it provides tools to build chemical intuition and directly evaluate the limits of the models. It is easy to visualize these 16 compounds and draw direct conclusions about the relationships between bond-types, location and source, without fully understanding the underlying machine learning classifier. The errors in this are also informative - illustrating the over-importance of internal single vs. double carbon bonds.

In this study we provide an application of Locally Interpetable Machine-Agnostic Explanations in the area of chemistry. Direct estimation of complex properties can be easily achieved using a variety of machine learning models, however tying the results of these models to specific structural features has been a challenge. We believe that the application of LIME to chemical structures provides an important advancement in machine learning for cheminformatics. This method has application for both evaluating {\it a priori} confidence in the machine learning models using to generate quantitative structure-functional predictions, as well as a tool for directing retrosynthesis efforts.

\subsubsection*{Acknowledgments}
\small
This research was conducted as part of the Co-Optimization of Fuels \& Engines (Co-Optima) Project sponsored by the Bioenergy Technologies and Vehicle Technologies Offices, Office of Energy Efficiency and Renewable Energy (EERE), U.S. Department of Energy (DOE). Co-Optima is a collaborative project of multiple national laboratories initiated to simultaneously accelerate the introduction of affordable, scalable, and sustainable biofuels and high-efficiency, low-emission vehicle engines. Sandia National Laboratories is a multi-program laboratory managed and operated by Sandia Corporation, a wholly owned subsidiary of Lockheed Martin Corporation, for the U.S. Department of Energy’s National Nuclear Security Administration under contract DE-AC04-94AL85000. This work was part of the DOE Joint BioEnergy Institute (http://www.jbei.org) supported by the U. S. Department of Energy, Office of Science, Office of Biological and Environmental Research, through contract DE-AC02- 05CH11231 between Lawrence Berkeley National Laboratory and the U.S. Department of Energy. The United States Government retains and the publisher, by accepting the article for publication, acknowledges that the United States Government retains a non-exclusive, paid-up, irrevocable, world-wide license to publish or reproduce the published form of this manuscript, or allow others to do so, for United States Government purposes. 

\section*{References}

\small
[1] Wang, Yanli, et al.\ (2009) "PubChem: a public information system for analyzing bioactivities of small molecules." {\it Nucleic acids research} {\bf 37}.suppl 2: W623-W633.

[2] Bohacek, Regine S., Colin McMartin, and Wayne C. Guida.\ (1996) The art and practice of structure-based drug design: A molecular modeling perspective. {\it Medicinal research reviews} {\bf 16}.1: 3-50.

[3] Whitmore, Leanne S., et al. (2016) BioCompoundML: a general biofuel property screening tool for biological molecules using Random Forest Classifiers." {\it Energy \& Fuels}.

[4] Ribeiro, Marco Tulio, Sameer Singh, and Carlos Guestrin. (2016) "Why Should I Trust You?": Explaining the Predictions of Any Classifier." {\it arXiv preprint} arXiv:1602.04938.

[5] Cheng, Tiejun, et al. (2007) Computation of octanol-water partition coefficients by guiding an additive model with knowledge. {\it Journal of chemical information and modeling} {\bf 47}.6 : 2140-2148.

[6] Pan, Yong, et al. (2008) Advantages of support vector machine in QSPR studies for predicting auto-ignition temperatures of organic compounds. {\it Chemometrics and Intelligent Laboratory Systems} {\bf 92}.2: 169-178.

[7] Bonchev, Danail. (2000) Overall connectivities/topological complexities: A new powerful tool for QSPR/QSAR. {\it Journal of chemical information and computer sciences} {\bf 40}.4: 934-941.

[8] de Lima Ribeiro, Fabiana Alves, and Márcia Miguel Castro Ferreira. (2003) QSPR models of boiling point, octanol–water partition coefficient and retention time index of polycyclic aromatic hydrocarbons. {\it Journal of Molecular Structure: THEOCHEM} {\bf 663}.1: 109-126.

[9] Organisation for Economic Co-Operation and Development. (2007) Principles for the Validation, for Regulatory Purposes, of (Quantitative) Structure–Activity Relationship Models. {\it OECD Environment Health and Safety Publications} {\bf 69}.

[10] Wang, Z., Liu, H., Song, T., Qi, Y., He, X., Shuai, S. and Wang, J. (2015) Relationship between super-knock and pre-ignition. {\it International Journal of Engine Research} {\bf 16}.2: 166-180.

[11] Theiss, T.J., Alleman, T., Brooker, A., Chupka, G., Elgowainy, M.A., Han, J., Huff, S.P., Johnson, C., Kass, M.D., Leiby, P.N. and Martinez, R. (2016) Summary of High-Octane Mid-Level Ethanol Blends Study {\it Fuels, Engines and Emissions Research Center} National Transportation Research Center (NTRC).

[12] Archer, K. J., Kimes, R. V. (2008). Empirical characterization of random forest variable importance measures. {\it Computational Statistics \& Data Analysis} {\bf 52}.4: 2249-2260.
\end{document}